# Iterative Nearest Neighborhood Oversampling in Semi-supervised Learning from Imbalanced Data

Fengqi Li, Chuang Yu, Nanhai Yang, Feng Xia*, Guangming Li, and

Fatemeh Kaveh-Yazdy

*School of Software, Dalian University of Technology, Dalian 116620, China*
*Corresponding author: Feng Xia; Email: f.xia@ieee.org*

**Abstract**
Transductive graph-based semi-supervised learning methods usually build an undirected graph utilizing both labeled and unlabeled samples as vertices. Those methods propagate label information of labeled samples to neighbors through their edges in order to get the predicted labels of unlabeled samples. Most popular semi-supervised learning approaches are sensitive to initial label distribution happened in imbalanced labeled datasets. The class boundary will be severely skewed by the majority classes in an imbalanced classification. In this paper, we proposed a simple and effective approach to alleviate the unfavorable influence of imbalance problem by iteratively selecting a few unlabeled samples and adding them into the minority classes to form a balanced labeled dataset for the learning methods afterwards. The experiments on UCI datasets and MNIST handwritten digits dataset showed that the proposed approach outperforms other existing state-of-art methods.

## 1 Introduction

In recent years, the booming information technology leads to databases included a massive amount of data in different fields. Subsequently, the need for mining useful potential is inevitable. The target classes of most of these data records, called unlabeled records, are unknown, and the records with specified target classes are called labeled records. Only a small ration of records are labeled because it is very time-consuming and labor-intensive to obtain annotates (labels) by domain experts. In machine learning, semi-supervised learning (SSL) methods [1] train a classifier by combining labeled and unlabeled samples together, which has attracted attentions due to their advantage of reducing the need for labeled samples and improving accuracy in comparison with most of supervised learning methods. However, although most existing methods have shown encouraging success in many applications, they assume that the distribution between classes in both labeled and unlabeled datasets are balanced, which may not satisfy the reality [2].

If the dataset only contains two classes, a binary classification, the class that has more samples is called the majority class, and the other one is called the minority class. Many popular SSL methods are sensitive to the initial labeled dataset and are suffered from a severe skew of data to the majority classes. In many real-world applications such as text classification [3], credit card fraud detection [4], intrusion detection [5], and classification of protein databases [6], datasets are imbalanced and skewed.

The imbalance learning problem [7] puzzles many machine learning methods established on the assumption that every class has the same or approximate same quantity of samples in raw data. There are various methods proposed to deal with the imbalance classification problems. These methods can be classified into re-sampling [8], cost-sensitive learning [23], kernel-based learning [24] and active learning methods [20, 25].

Re-sampling methods include oversampling [19, 21] and undersampling [14] approaches, in which the class distribution is balanced by adding a few of samples to minority class or removing a few of samples from majority class, respectively.

Most existing studies on imbalanced classification focus on supervised imbalanced classification instances [8, 14, 19, 23], and there are few studies on semi-supervised methods for imbalanced classification [4]. The bias caused by differing class balances can be systematically adjusted by re-weighting [15, 16] or re-sampling [17].

Focused on the bad performance of SSL algorithm to the imbalanced learning problem, we propose a novel approach based on oversampling in consideration of the SSL's characteristic that there are abounds of unlabeled samples. Li et al. combined active learning with SSL methods that sample a few of most helpful modules for learning a prediction model in [20]. Based on above considerations, **I**terative **N**earest **N**eighborhood **O**versampling (INNO) algorithm we propose in this paper tries to convert a few unlabeled samples to labeled samples for minority classes, consequently constructing a balanced or approximately balanced labeled dataset for standard graph-based SSL methods afterwards. Therefore, we aim to alleviate the unfavorable impact of typical classifiers in dealing with imbalanced dataset in SSL domain.

In this paper, we provide an effective and efficient heuristic method to eliminate the 'injustice' brought by imbalanced labeled dataset. As the samples with a close affinity in a low dimension feature space will probably have the same label, we propose an iterative search approach to simply oversample a few unlabeled samples around known labeled samples in order to form a balanced labeled dataset. Extensive experiments on synthetic and real datasets confirm the effectiveness and efficiency of our proposed algorithms.

The remainder of this paper is organized as follows. In Section 2, we provide a brief review of existing studies of semi-supervised learning and their applications on imbalanced problem. We give the motivation behind the proposed INNO in Section 3. In Section 4, we revisit some popular algorithms by giving a graph transduction regularization framework, and then we introduce our proposed algorithm INNO in details. The experimental results on some imbalanced dataset are presented in Section 5. Finally, we conclude the paper in Section 6.

## 2 Related Work

As SSL accomplishes an inspiring performance in combining a small scale of labeled samples and a mass mount of unlabeled samples effectively, it has been utilized in many real-world applications such as topic detection, multimedia information identification and object recognition. For the past few years, graph-based SSL approaches have attracted increasing attention due to their good performance and ease of implementation. Graph-based SSL regards both labeled and unlabeled samples as vertices in a graph and builds edges between pairwise vertices, and the weight of edge represents the similarity between the corresponding vertices. Transductive graph-based SSL methods predict the label for unlabeled samples via graph partition or label propagation using a small portion of seed labels provided by initial labeled dataset [22]. Popular transductive algorithms include the Gaussian fields and harmonic function based method (GFHF) [10], the local and global consistency method (LGC) [11],the graph transduction via alternating minimization (GTAM) [15], popular inductive methods consist of transductive support vector machines (TSVM) and manifold regularization [9]. Recent researches on graph-based SSL include ensemble manifold regularization [18] and relevance feedback [12]. However, these graph-based SSL methods developed with smoothness, clustering assumption, and manifold assumption [1] frequently perform a bad classification if provided an imbalanced dataset.

Wang et al. [15] proposed a node regularizer to balance the inequitable influence of labels from different classes, which can be regarded as a re-weighting method. They developed an alternating minimization procedure to interleave optimize the node regularizer and classification function, and greedily searched the largest negative gradient of cost function to determinate the label of an unlabeled samples during each minimization step until acquiring all predicted labels of unlabeled samples. Nevertheless, the time complexity of the algorithm is $O(n^3)$, and also it would be suffered from error occured in classification progress, during iteration. Its modified algorithm LDST [16] revises the unilateral greedy search strategy into a bidirectional manner, which can drive wrong label correction in addition to eliminate imbalance problem.

Other graph-based SSL algorithms solve imbalance problem mainly by re-sampling methods. Li et al. [2] proposed semi-supervised learning with dynamic subspace generation algorithm based on undersampling to handle imbalanced classification. They constructed several subspace classifiers on the corresponding balanced subset by iteratively performing under-sampling without duplication on majority class to form a balanced subnet. However, the algorithm features high complexity in computational time.

## 3 Motivation

Transductive graph-based SSL methods propagate label information of labeled samples to their neighbors through edges to get the predicted labels of unlabeled samples. Once there is an imbalanced distribution of classes in labeled dataset, the class boundary will severely skew to the majority classes, which have a more possibility to influence the predicted labels of unlabeled samples. We draw the influence of imbalance classification result to three popular transductive GSSL methods on the two-moon toy dataset in Figure 1. The symbols '□' and '▽' stand for class '+1' and '-1' respectively in raw data, and we use solid symbol to depict labeled data. Originally, class '+1' contains one labeled samples and class '-1' contains ten labeled data. In Figure 1, it can be seen that the impact of imbalance label distribution to aforementioned algorithms even on a well-separated dataset. The conventional transductive graph-based SSL algorithms, such as GFHF [10], LGC [11], and GTAM [15], fail to give the acceptable classification result.

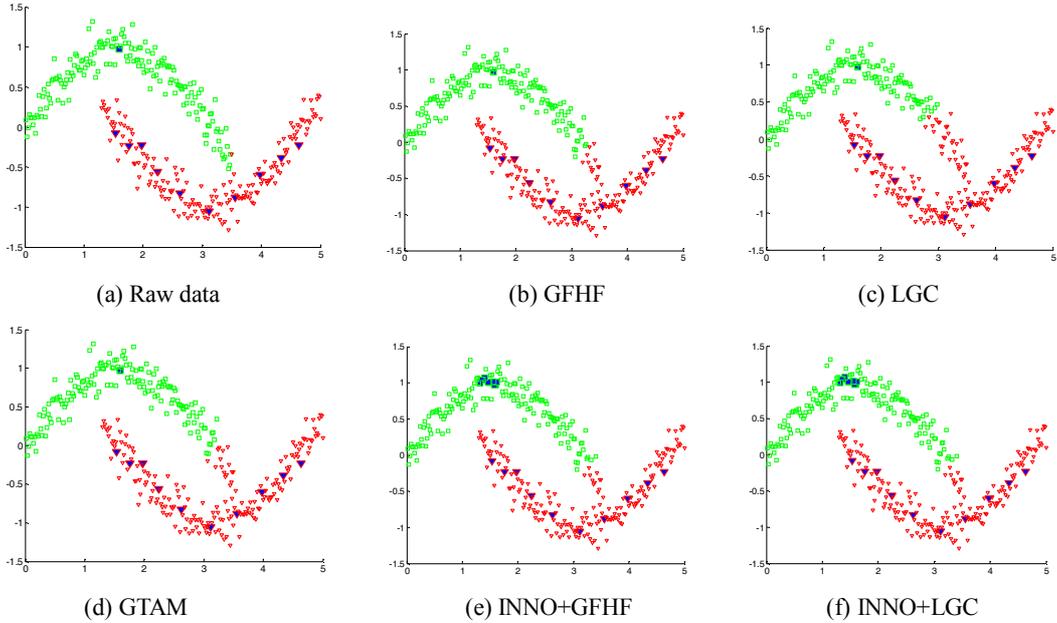

Figure 1. A demonstration of imbalanced label dataset affection to transductive GSSL methods on two-moon toy dataset.

Oversampling methods have been shown to be very successful in handling with imbalanced problem. However, Barua et al. [19] reported some cases of insufficiencies and inappropriateness in existing methods. They proposed MWMOTE that generated synthetic minority samples by using clustering approach to select samples according to data importance around a subnet of the minority class; however it achieves to select minority samples around the class boundary under a large number of training set. Plessis and Sugiyama [17] proposed a semi-supervised learning method to estimate the class ratio for test dataset by combining train and test dataset in supervised learning. However, these methods are inapplicable in SSL scenario.

In order to handle with the imbalance problem of labeled dataset in SSL scenario, considering the problem of abundant unlabeled samples in SSL domain, we proposed a simple and effective method, called Iterative Nearest Neighborhood Oversampling, to convert a few of unlabeled samples to labeled samples for minority class, which can construct a balanced labeled samples for learning methods. We integrate the proposed algorithm with two popular transductive graph-based SSL methods to perform a robust classification to imbalanced problem, and the processing flow can be described as Figure 2.

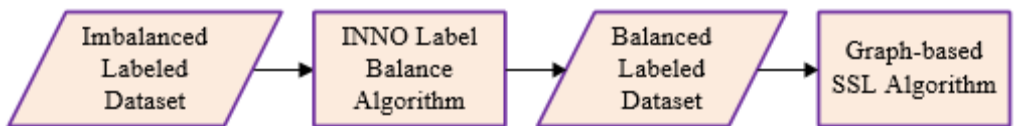

Figure 2. Workflow of graph-based SSL integrating with INNO

## 4 Iterative Nearest Neighborhood Oversampling
### 4.1 Graph-based SSL Formulation
Given a raw dataset $X = \{X_L \cup X_U\}$ containing $n$ samples, where $X_L = \{(x_1,y_1), (x_2,y_2),\ldots,(x_l,y_l)\}$ is the labeled dataset with cardinality $|X_L| = l$ and $X_U = \{x_{l+1}, x_{l+2},..., x_{l+u}\}$ is the unlabeled dataset with cardinality $|X_U| = u$, where $l+u = n$ and typically $l << u$. Define the labels correlation to labeled dataset is $Y_L = \{y_1, y_2,..., y_l\}$, where $y_i \in \{1,2,...,c\}(i=1...l)$ and $c$ is the count of different classes. The goal of SSL is to infer $Y_U = \{y_{l+1}, y_{l+2},..., y_{l+u}\}$ of unlabeled samples by combining $X_L$ and $X_U$.

Graph-based SSL formulate an undirected graph $G = \{X, E, W\}$ in which the vertex set is $X = \{x_i \in R^d\}$ ($i=1...n$, $d$ is the number of features) and the edge set between vertex is $E = \{e_{ij}\}$. The samples are treated as vertex and the edges $e_{ij}$ can be weighted by $W_{ij} = k(x_i, x_j)$, where $k(x_i, x_j)$ could be a similarity measure such as Euclidean distance, RBF distance or cosine distance; thus, the weight matrix can be represented as $W = \{W_{ij}\}$.

Define the graph Laplacian $\Delta = D - W$ and the normalized graph Laplacian is $L = D^{-1/2}\Delta D^{-1/2} = I - D^{-1/2}WD^{-1/2}$, where $D = diag\{D_{11}, D_{22},..., D_{nn}\}$ is the node degree matrix with diagonal element $D_{ii} = \sum_j W_{ij}$. The binary label matrix $Y = \{Y_{ij} \in B^{n \times c}\}$ is set as $Y_{ij} = 1$ if $x_i$ is labeled as class $j$ and $Y_{ij} = 0$ otherwise.

Most Graph-based SSL methods perform label propagation procedure based on manifold assumption, that is, the labels are smooth on the graph. Therefore, they essentially estimate a classification function $\{F: X \to R^{n \times c}\}$ constrained to give the true label for labeled samples and given smooth labels over the whole graph. Mathematically, Graph-based SSL methods formulate a regularization framework by a cost function as follows:
$$Q\{F\} = Q_l(F) + Q_s(F) \qquad (1)$$
where $Q_l(F)$ is a loss function to penalize the deviation from the given labels, and $Q_s(F)$ is regarded as the smooth regularizer to prefer the label smoothness. The optimal $F^* = \arg\min_{F \in \mathbb{F}} Q(F)$ can be calculated by minimization the cost function $Q\{F\}$. Therefore, different Graph-based SSL methods can be obtained by assigning different loss functions and regularizers to $Q_l(F)$ and $Q_s(F)$.

**4.2 Methodology**
In real-world applications, labeled samples are always sampled according with normalization distribution. Labels of samples in some classes are easy to obtain, and in others that are not, even if they are of the same important level. To deal with the imbalance classification problem in semi-supervised learning scenarios, we assume that there are lots of unlabeled samples around a labeled sample in a low feature dimension space. Therefore, we can select a few unlabeled samples for minority class to form a balanced dataset. We describe our oversampling model as follows:

Consider the multi-class classification scenarios, let $r = \{r_1, r_2,..., r_c\}$ denotes the size set of labeled samples in labeled dataset, where $r_j (j = 1...c)$ is the number of labeled samples in class $j$. We use standard variance $var(r)$ represent the dispersion degree of the quantity of labeled samples in each class, and the imbalance ratio $var(r)$ can be describe as follows:
$$var(r) = \left(\frac{1}{c}\sum_{j=1}^{c}(r_j - \bar{r})^2\right)^{1/2} \qquad (2)$$
where $\bar{r} = \frac{1}{c}r_{sum}$, $r_{sum} = \sum_{j=1}^{c}r_j$.

We propose a novel approach to iteratively increase the labeled samples of the minority class, named **I**terative **N**earest **N**eighborhood **O**versampling (INNO), in order to eliminate the adverse influence of imbalanced labeled dataset. During iteration, we obtain the class $j$ containing the smallest number of labeled data, and traverse $k$ neighbors of each labeled samples in class $j$ to select the most similar sample to all labeled samples of class $j$ in the unlabeled dataset. The most similar sample can be defined as:
$$x_{\max k} = \arg\max_{x_k \in X_u} k(x_k, x_j) \qquad (3)$$
where $x_j$ is the labeled samples in class $j$.

To avoid $x_{\max k}$ deriving from classification boundary, we skip the samples which are connected to labeled samples of remainder classes. Then, we simply label the sample $x_{\max k}$ with class $j$, remove it from unlabeled dataset $X_U$ and add it to the labeled dataset $X_L$. We formalize the INNO approach as algorithm 1.

**Algorithm 1** Iterative Nearest Neighborhood Oversampling (INNO)
**Input:** $k$ NN graph, affinity matrix $W$, stop parameter $s$, imbalanced labeled dataset $X_L$ and unlabeled dataset $X_U$;
**Output:** balanced or approximate balanced labeled dataset.
**Procedure:**
1  while var($r$) > $s$
2    Initialization $r_j = \min_{j \in 1...c} r$, max = $-\infty$, max$k$ = 0;
3    for each labeled sample $x_j$ in class $j$
4       for each neighbors $x_k$ of $x_j$
5          skip the $x_k$ if it is in $X_L$ or $x_k$ has edges between labeled samples in other class;
6          if $W_{ij}$ > max, then update max, max$k$
7       end for
8    end for
9    if max$k$=0 // all the neighbors of labeled samples in class $j$ have edges with labeled samples in other classes, then $r_j = r_{max}$, continue;
10   label $x_{maxk}$ with class $j$, remove it from $X_U$, add it to $X_L$, $r_j = r_j + 1$;
11 end while

As labeled dataset is very scarce scale compared with unlabeled samplesset in the background of semi-supervised learning, it's difficult to infer the class boundary by a small number of labeled data, caused by intrinsic sample selection bias or inevitable non-stationarity. Therefore, classic oversampling methods [19, 23] is not capable in this situation, because they need to judge of the informative data close to class boundary, in order to synthetically generate new samples for minority class. In contrary, we try to skip the unlabeled samples close to class boundary to reduce the risk of introducing reckless mistakes in SSL scenarios. So, we simply set $r_j = r_{max}$ if the iteration finds all the neighbors of labeled samples in class $j$ have edges with labeled samples in other classes, that is, no more samples will be introduced for class $j$. Moreover, our method is capable of multi-class classification, though most sampling methods are used to diagnose between-class imbalance problem.

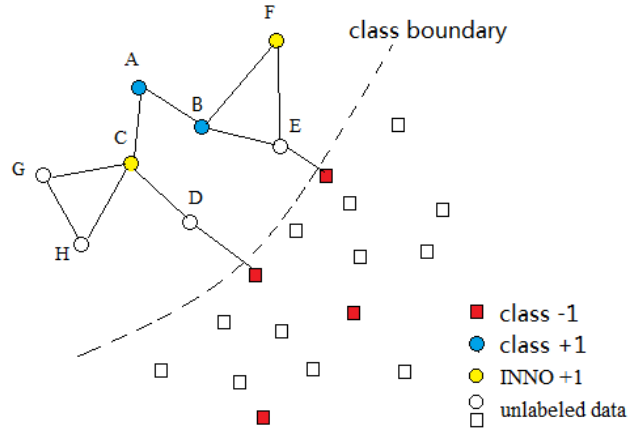

Figure 3. INNO algorithm illustration

Here we consider a binary classification demonstration in Figure 3, where the stars and circles represent the samples of majority and minority class respectively and the yellow points are unlabeled samples. The imbalance ratio of labeled dataset between class '+1' and '-1' is $r_{+1}:r_{-1}$ = 2:4. We employ a $k$-nearest neighbor classifier on the graph (assuming $k$ = 2) and only consider the neighborhood connections in class '+1'. We set the stop parameter $s$ = 0, that is, the iteration will stop when all classes have the same quantity of samples. As we can see, sample A and B are the initial labeled samples in minority class '+1', and then we show the process of INNO algorithm to balance the labeled dataset. The algorithm searches all neighbor unlabeled samples of A and B, finding the closest sample C which is not in labeled dataset and have no connections to labeled samples of class '-1', therefore, label C with '+1' and remove it from unlabeled dataset and add it into labeled dataset. The algorithm continues to search the neighbors of A, B and C to find the sample D, but the D is connected to labeled sample of class '-1', so it skips D and E as well. Thereby it finds sample F which satisfies all search conditions. At this moment, a balanced labeled dataset is obtained, and the algorithm ends with $s = 0$.

### 4.3 Complexity analysis

Our method query k neighbors of every labeled sample in each iteration, the time of query is $(r_{sum} + r_{sum} \times k) \times k$, and the time of iteration in the worst situation is $r_{max} \times c - r_{min} \times (c-1)$, where $r_{max}$ and $r_{min}$ is the largest and smallest number of labels. The time complexity of the proposed algorithm is $O(c \times r_{max} \times k^2 \times \frac{r_{max}(r_{max}-1)}{2} \times k) = O(ck^3 r_{max}^3)$. As the scale of labeled samples is small, thus the $r_{max} < l << n$ inequality is held. Clearly, class number $c$ and neighbor number $k$ are very small, thus resulting in low computational complexity for our algorithm.

## 5 Experiments

There are many accuracy measures for evaluating the two-class classification problems, such as precision, recall, geometric-mean (G-mean) and F-measure [12]. To evaluate the classifier performance, we calculate the accuracy by a confusion matrix as illustrated in Table 1.

Table 1. Confusion matrix

|  | True class | |
|---|---|---|
| Y | TP (True Positives) | FP (False Positives) |
| N | FN (False Negatives) | TN (True Negatives) |
| Row sum | p | n |

According to Table 1, many performance measures can be derived and domain classes are regarded as positive and negative classes. One of the most common criteria is overall accuracy which is used for two class classification problems in this paper.

It can be defined as

$$accuracy = \frac{TP + TN}{p + n} \quad (4)$$

This measure provides a simple way of describing a classifier's performance on a given data set. Meanwhile, we apply RBF kernel function $W_{ij} = \exp(\Sigma_d |x_{id} - x_{jd}|^2/\sigma^2)$ to calculate the similarity measure between samples, and set parameter $\alpha=0.99$ in LGC [9] and $\mu = 99$ in GTAM [14] during the experiments and the results are the average results of 50 runs. All the experiments are run on a Dell Optiplex-380 PC with Intel Pentium dual-core processors 2.93GHz and main memory of 3GB.

### 5.1 UCI datasets

Firstly we evaluate the effectiveness of our proposed INNO algorithm combined with SSL methods on IRIS and IONOSPHER datasets from UCI repository. IRIS dataset consists of three different categories of flower, 'setosa', 'versicolor' and 'virginica'. Each category contains 50 samples, and the feature dimension of a sample is 4. We fix the number of labeled samples in category 'setosa' at 10, but range the number of labeled samples in category 'versicolor' from 1 to 10, and also range the number of labeled samples in category 'virginica' from 10 to 20. Stop parameter $s$ is set to zero on this dataset, namely the balance algorithm stops when the labeled dataset are completely balanced. Set $\sigma = 0.26$ in RBF kernel function and the neighbors $k = 5$ in $k$-NN. Figure 5(a) shows the classification result on IRIS.

IONOSPHERE dataset has 351samples of 'g' and 'b' categories. Category 'g' and 'b' contains 225, 126 respectively, and each sample has 34 dimensions. We set the stop parameter $s=0$ in INNO algorithm. As the class distribution in the original data set are not balanced, so that the label balance algorithm can stop at the right point where the imbalance ratio of labeled dataset are consistent with the original dataset. Zhu et al. proposed in CMN method in [10] to solve the negative influence to the classification result caused by imbalanced labeled dataset. We compared it with our algorithm in this paper. In this experiment, we set the size of labeled samples in category 'g' by ranging from 12 to 21 and set the size of labeled samples in category 'b' range of 2~11. Set $\sigma = 1$ in RBF kernel function and $k = 10$ in $k$-NN. Figure 5(b) shows the classification accuracy on IONOSPHERE.

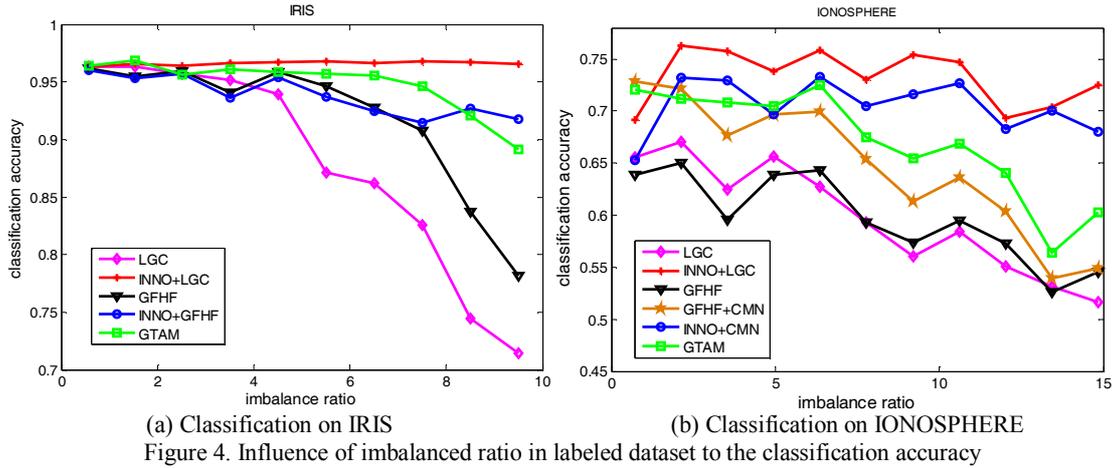

(a) Classification on IRIS  (b) Classification on IONOSPHERE
Figure 4. Influence of imbalaced ratio in labeled dataset to the classification accuracy

It can be seen in Figure 4(a) that all algorithms have high classification accuracy when each class has the same number of labeled data. As the imbalanced ratio increases between different classes in labeled dataset, classification accuracy drops gradually in GFHF, LGC and GTAM algorithm, around 70% when imbalance ratio is about 9, while the proposed INNO+GFHF and INNO+LGC algorithms remain stable, basically maintained about 95%. Therefore, INNO algorithm shows a better robustness when dealing with imbalanced labeled dataset. We can get similar results from Figure 4(b), although GRF+CMN method can reduce the influence of imbalanced labeled dataset to classification results to same extent, it's under the assumption that the labeled dataset have the same distribution with the original data from the class definitely. Therefore, the classification accuracy drops when the class distribution in the original data is different from the class distribution in labeled dataset.

## 5.2 Handwritten digit dataset

In this section, we conduct two classification experiments on MNIST handwritten digit dataset. MNIST handwritten dataset has a training set of 60000 samples and a test set of 10000 samples, each sample has a pixel of 28×28, and each pixel is a grayscale range from 0 to 255. In these experiments, we combined the training and test set together and the pixel values of the image were used directly as features, i.e. each sample has a feature number of 784. We randomly selected 200 samples of the number '0' to '9' from the entire data set, so the sample set has 2000 samples. We used the parameter σ = 380 that Zhu et al. [8] set in MNIST and set the stop parameter $s = 0$ in INNO. We selected digit '5' to '9' to conduct a 5-class experiment and digit '0' to '9' to conduct a 10-class experiment. Figure 5 shows that the classification accuracy of each algorithm curves as we continue increasing the imbalanced ratio of labeled dataset.

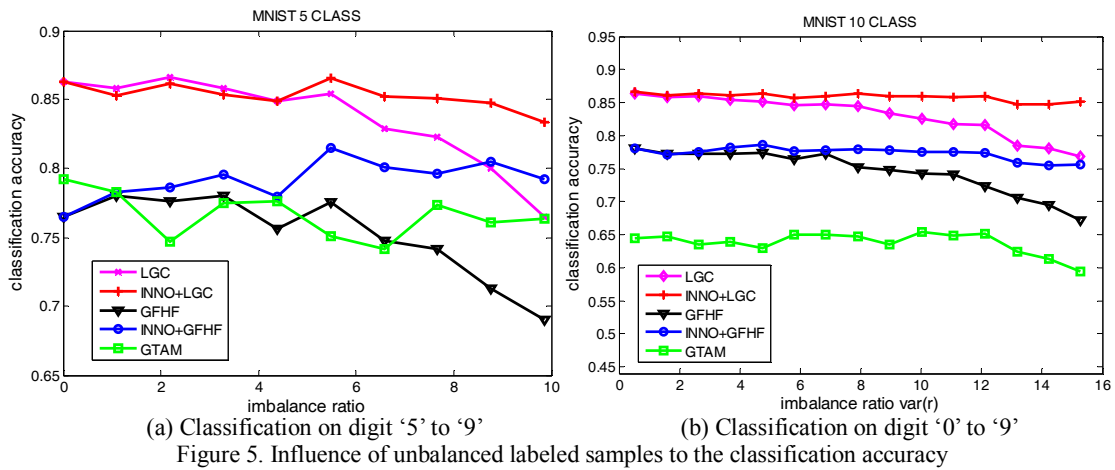

(a) Classification on digit '5' to '9'  (b) Classification on digit '0' to '9'
Figure 5. Influence of unbalanced labeled samples to the classification accuracy

Figure 5 illustrates the label balance algorithm is not necessary when the imbalanced ratio approximates to 0, therefore, LGC ,INNO+LGC, GFHF and INNO+GFHF algorithm have the same classification accuracy, namely the algorithm we propose will not affect the accuracy of the original algorithm when the labeled dataset is balanced at the beginning. While the classification accuracy of GFHF, LGC and GTAM is decreased significantly when the imbalance ratio increases gradually, INNO+GFHF and INNO+LGC raised in the figure show a stable performance. It also can be seen that the experiments on digit '5' to '9' and digit '0' to '9' show that the accuracy of GTAM algorithm decreases along with the class numbers increasing obviously, while others are not.

**5.3 Parameter discussion**

Intuitively, the number of neighbors, $k$, will affect the result of INNO algorithm. To validate this conjecture, we perform an experiment on the UCI datasets. We fix the imbalanced ratio of labeled dataset at 10:1:20 with categories 'setosa', 'versicolor' and 'virginica' on IRIS dataset, so $var(r)$ = 9.50, and fix the imbalanced ratio of labeled dataset at 23:2 between class 'g' and 'b' on IONOSPHERE dataset, so $var(r)$ = 14.85. Set the stop parameter $s$ to zero, the classification accuracy trends are shown in Figure 6.

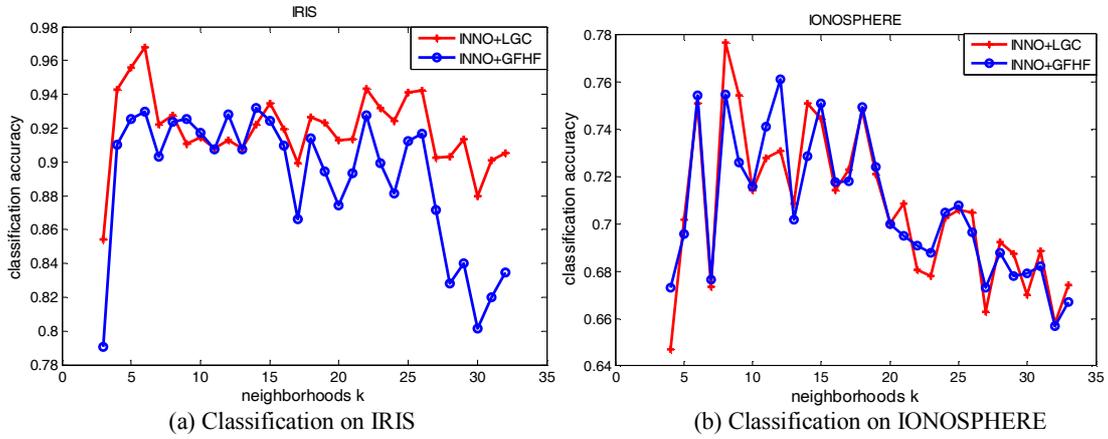

(a) Classification on IRIS  (b) Classification on IONOSPHERE
Figure 6. Influence of neighbors $k$ to classification accuracy

The classification accuracy is not high when $k$ value is too small, so by increasing $k$, the classification accuracy increases drastically, and then the classification accuracy on IRIS fluctuates lenitively around 90%~95% when $k$ remains in a certain region. However, if $k$ continues to increase, the classification accuracy begins to drop severely down to 75%. It's because when $k$ is too large, the number of nearest neighbors is excessive, so INNO algorithm will find that all the neighbors are connected to other categories. Then INNO algorithm is unable to balance the labeled samples, resulting in lower classification accuracy.

Moreover, we consider the influence of stop parameter $s$ to classification accuracy. We perform another experiment on IRIS and IONOSPHERE with $k$ to 5 and 10 respectively and fix the imbalanced ratio as same as above. We change the stop parameter $s$ to observe the oscillation on classification accuracy in Figure 7. As we can see from Figure 7(a), it shows little improvement of the classification accuracy when the stop parameter and the original imbalanced ratio are nearly the same. At this point, INNO algorithm doesn't convert enough unlabeled samples into labeled data. The labeled dataset tends to become more and more balanced and the classification accuracy increases quickly and falls in a certain range, when the step parameter is decreased.

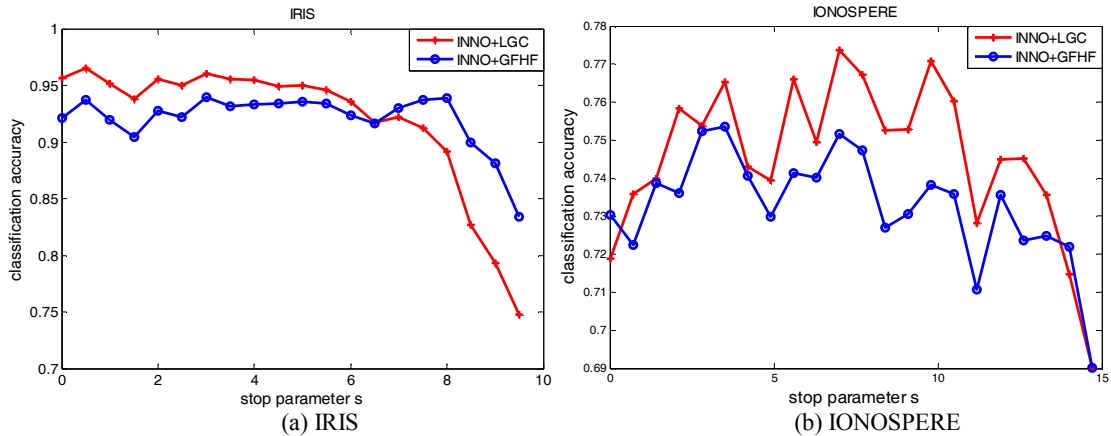

(a) IRIS  (b) IONOSPERE
Figure 7. Influence of stop parameter *s* to classification accuracy

Improved classification accuracy could be achieved by choosing smaller stop condition values. But the results are not as supposed and oscillate in a certain range, as shown in Figure 7(b). One possible reason is that the INNO algorithm skips the classification boundary of unlabeled samples at algorithm step 5 if all the neighbors of labeled samples in minority class have edges with labeled samples in other classes. When the stop parameter becomes smaller, the more unlabeled samples will be selected and the probability of the unlabeled samples close to classification boundary is higher. As the algorithm searched the neighbors of labeled samples in minority class, the labeled number of the current class will be simply set to $r_{max}$ at the algorithm step 9; therefore, it will not introduce new labeled samples in this class. And as well taking into account of the randomness of labeled samples selected from raw data by algorithm, we cannot foresee the occurrence of such a situation [17], namely the imbalanced output may also occur even the stop condition *s* is 0, so the classification accuracy ranges in a certain scope.

## 6 Conclusion

In classification scenarios, state-of-art semi-supervised learning methods estimate a classification function on the assumption that there is a balanced distribution in labeled and unlabeled dataset. However, the class boundary will be severely skewed by majority class in an imbalance between-class are, which is proved by the experiments on UCI datasets and MNIST digit recognition.

As the bias caused by disproportionately imbalanced dataset adjusted by re-sampling or re-weighting, we proposed the INNO algorithm to settle down this imbalance problem simply and effectively, which eliminates the 'injustice' brought by imbalanced labeled dataset to popular transductive graph-based SSL methods. Our method iteratively searches the neighbors of labeled samples in minority class to seek out the nearest neighbor to all labeled samples of minority class, and try to skip the unlabeled samples close to the class boundary. Therefore, we can construct a balanced or approximately balanced labeled dataset for the learning methods. The experiments show a better classification result of SSL methods combined with INNO.